\documentclass[letterpaper]{article} 

\usepackage{aaai2026}  
\usepackage{times}  
\usepackage{helvet}  
\usepackage{courier}  
\usepackage[hyphens]{url}  
\usepackage{graphicx} 
\usepackage{natbib}  
\usepackage{marvosym} 
\usepackage{caption} 
\usepackage{algorithm}
\usepackage{algorithmic}
\usepackage{amsmath}
\usepackage{amssymb}
\usepackage{booktabs}
\usepackage{multirow}
\usepackage[table]{xcolor}
\usepackage{makecell}
\usepackage{subcaption}
\usepackage[switch]{lineno}
\usepackage{newfloat}
\usepackage{listings}
\DeclareCaptionStyle{ruled}{labelfont=normalfont,labelsep=colon,strut=off} 
\floatstyle{ruled}
\newfloat{listing}{tb}{lst}{}
\floatname{listing}{Listing}
\title{MEML-GRPO: Heterogeneous Multi-Expert Mutual Learning for RLVR Advancement}
\author{
Weitao Jia\textsuperscript{\rm 1}* , Jinghui Lu\textsuperscript{\rm 1}*\dag , Haiyang Yu\textsuperscript{\rm 1 2}*, Siqi Wang\textsuperscript{\rm 1}, Guozhi Tang\textsuperscript{\rm 1}, An-Lan Wang\textsuperscript{\rm 1}, Weijie Yin\textsuperscript{\rm 1}, Dingkang Yang\textsuperscript{\rm 1}, Yuxiang Nie\textsuperscript{\rm 1}, Bin Shan\textsuperscript{\rm 1}, Hao Feng\textsuperscript{\rm 1}, Irene Li\textsuperscript{\rm 3}, Kun Yang\textsuperscript{\rm 2}, Han Wang\textsuperscript{\rm 1}, Jingqun Tang\textsuperscript{\rm 1}, Teng Fu\textsuperscript{\rm 2}, Changhong Jin\textsuperscript{\rm 4}, Chao Feng\textsuperscript{\rm 1}\dag, Xiaohui Lv\textsuperscript{\rm 1}\dag, Can Huang\textsuperscript{\rm 1}\dag
}
\affiliations{
    \textsuperscript{\rm 1} ByteDance Inc.\\ 
    \textsuperscript{\rm 2} Fudan University\\
    \textsuperscript{\rm 3} University of Tokyo \hspace{1em} \\
    \textsuperscript{\rm 4} University College Dublin\\

}

\usepackage{bibentry}

\begin{document}

\maketitle
\vspace*{20pt}
\begingroup
\renewcommand\thefootnote{}%
\footnotetext{* Equal contribution.}%
\footnotetext{\dag\ Corresponding author.}%
\endgroup

\begin{abstract}
Recent advances demonstrate that reinforcement learning with verifiable rewards (RLVR) significantly enhances the reasoning capabilities of large language models (LLMs). However, standard RLVR faces challenges with reward sparsity, where zero rewards from consistently incorrect candidate answers provide no learning signal, particularly in challenging tasks. To address this,we propose \textbf{M}ulti-\textbf{E}xpert \textbf{M}utual \textbf{L}earning \textbf{GRPO} (MEML-GRPO), an innovative framework that utilizes diverse expert prompts as system prompts to generate a broader range of responses, substantially increasing the likelihood of identifying correct solutions. Additionally, we introduce an inter-expert mutual learning mechanism that facilitates knowledge sharing and transfer among experts, further boosting the model’s performance through RLVR. Extensive experiments across multiple reasoning benchmarks show that MEML-GRPO delivers significant improvements, achieving an average performance gain of 4.89\% with Qwen and 11.33\% with Llama, effectively overcoming the core limitations of traditional RLVR methods.
\end{abstract}

\section{Introduction}

Recent advances in large language model (LLM) and reasoning~\cite{kojima2022large,wei2022chain,wang2022self,lyu2023faithful,feng2023towards,shah2024causal,zhang2024chain,yu2023chinese,wang2025eliciting,zhang2025llms,zhou2025core,lu-etal-2025-bounding,lu2025prolonged,yu2025eve,lu2024padellm,lu2023punifiedner,lu-etal-2023-makes,lu-etal-2022-rationale} have showcased the efficacy of reinforcement learning with verifiable rewards (RLVR) in improving reasoning capabilities. Models such as OpenAI-o1~\cite{jaech2024openai}, DeepSeek-r1~\cite{shao2024deepseekmath,guo2025deepseek}, Doubao-1.5-thinking~\cite{bytedance_seed_thinking_2025}, and Qwen QwQ~\cite{team2024qwq} demonstrate the effectiveness of this approach in enhancing logical and problem-solving performance. By optimizing models based on binary correctness signals---such as matching ground truth solutions~\cite{xia2025evaluating} in mathematics or passing unit tests in code~\cite{lai2025analogcoder}---RLVR provides a scalable and automated approach to improving reasoning without relying on expensive human annotations~\cite{ouyang2022training,tong2024cambrian} or Process Reward Models~\cite{setlur2024rewarding}. 

Despite its successes, some studies~\cite{yue2025does,shojaee2025illusion,zhao2025echo} argue that RLVR does not endow LLMs with genuinely new reasoning abilities beyond those already present in their base models. Instead, RLVR primarily improves performance by steering models toward reasoning paths they already know, which are more likely to yield rewards. In other words, most RLVR methods optimize within the model’s existing knowledge rather than enabling the acquisition of new information. This limitation is most evident in what is termed \textit{reward sparsity}: when a model’s initial policy fails to produce correct responses in complex reasoning tasks, the lack of positive learning signals prevents RLVR from driving meaningful progress. This limitation becomes particularly pronounced in scenarios requiring exploration beyond the model’s current knowledge. The on-policy nature of current RLVR methods, which rely solely on a model’s own generated trajectories for learning, inherently limits their ability to explore beyond existing capabilities. When a model’s initial reasoning capacity is inadequate, standard RLVR tends to plateau early. This constraint, closely tied to the issue of reward sparsity described earlier, highlights a critical challenge: \textit{How can RLVR be designed to overcome reward sparsity and uncover correct reasoning paths, even when the initial policy falls short?}

To overcome this challenge, we introduce Heterogeneous \textbf{M}ulti-\textbf{E}xpert \textbf{M}utual \textbf{L}earning \textbf{GRPO} (MEML-GRPO), an innovative framework that leverages the complementary strengths of multiple heterogeneous pre-trained models to overcome performance bottlenecks. Our approach is inspired by the observation that diverse, heterogeneous reasoning models often generate varied and complementary solutions to the same problem. As shown in Table~\ref{tab:error_analysis}, error distributions in GSM8K~\cite{cobbe2021training} across these models (\textit{i.e.,} DeepSeek-r1, GPT4o\footnote{Model card:gpt-4o-2024-05-13} and Doubao-1.5-thinking) exhibit low overlap, with only 3.06\% of errors shared among all three models. Notably, in the GSM8K dataset, 90.11\% of incorrect predictions from GPT4o can be corrected by other heterogeneous models, highlighting the potential for mutual learning to enhance overall performance.

\begin{table*}[t]
\centering
\footnotesize
\scalebox{0.8}{
\begin{tabular}{l|cccc}

\toprule
\textbf{Metric} & \textbf{DeepSeek-r1} & \textbf{GPT4o} & \textbf{Doubao-1.5-thinking} & \textbf{Error Overlap} \\ \midrule
\textbf{GSM8K} & & & & \\ \midrule
Total errors & 748~(10.0\%) & 2316~(30.9\%) & 388~(5.1\%) & 229~(3.0\%) \\ \midrule
Errors corrected by other models & 519~(69.3\%) & 2087~(90.1\%) & 159~(40.9\%) & -- \\ \midrule
\textbf{StrategyQA} & & & & \\ \midrule
Total errors & 400~(20.0\%) & 326~(16.3\%) & 369~(18.4\%) & 192~(9.6\%) \\ \midrule
Errors corrected by other models & 208~(52.0\%) & 134~(41.1\%) & 177~(47.9\%) & -- \\ \midrule
\textbf{MathQA} & & & & \\ \midrule
Total errors & 3893~(13.0\%) & 5004~(16.7\%) & 3164~(10.6\%) & 2281~(7.6\%) \\ \midrule
Errors corrected by other models & 1612~(41.4\%) & 2723~(54.4\%) & 863~(27.9\%) & -- \\ \midrule
\end{tabular}}
\caption{Error distribution analysis across heterogeneous models on different datasets. Percentages show error rates (total errors/examples) and correction rates (errors fixed by other models/total errors). Error overlap indicates shared errors.}
\label{tab:error_analysis}
\end{table*}

By fine-tuning the policy model with responses from diverse models, MEML-GRPO boosts the likelihood of producing at least one valid reasoning path, delivering the essential learning signal for RLVR. Moreover, instead of merely aggregating the learned outputs of the policy model from these heterogeneous models, MEML-GRPO enables these models to learn from each other's strengths and promote mutual improvement. MEML-GRPO introduces three key innovations to enhance reasoning in RLVR:

\begin{enumerate}
    \item \textbf{Multi-Expert Fine-Tuning (MEF)} (Section~\ref{sec:mef}): Instead of relying on a single reasoning dataset, MEML-GRPO utilizes multiple system prompts, referred to as \textit{experts} in this work, each emulating distinct reasoning paths from diverse external models. These experts are fine-tuned on responses generated by heterogeneous models, with their unique reasoning styles captured through tailored system prompts, which bring a diverse set of reasonings that enrich the learning process.

    \item \textbf{Reinforced Inter-Expert Learning (RIEL)} (Section~\ref{sec:riel}): Beyond independent exploration, MEML-GRPO enables experts to learn from one another’s successful reasoning paths through a shared mechanism. Weaker experts improve by learning high-reward trajectories from stronger ones. As a result, all experts can perform competitively majority voting during inference.

    \item \textbf{Hard Example Accumulation via SFT Buffer} (Section~\ref{sec:riel}): When all experts fail to solve a problem, MEML-GRPO defaults to supervised fine-tuning (SFT) using ground truth, ensuring non-zero gradients for learning, maintaining progress even on challenging problems.

\end{enumerate}

MEML-GRPO strikes an optimal balance between exploration and exploitation. Unlike traditional RLVR methods that may stall on difficult problems, our framework dynamically integrates complementary strengths from multiple reasoning strategies, ensuring steady learning progress. Our contributions are:

\begin{itemize}
    \item We propose MEML-GRPO, which leverages system prompts to capture diverse reasoning styles and introduces a reinforced mutual learning algorithm for cross-expert knowledge transfer.
    \item Extensive experiments on GSM8K, MathQA, and StrategyQA demonstrate consistent performance improvements of 3.08\%, 8.65\%, and 2.96\% with Qwen and 5.87\%, 16.78\%, and 11.35\%  with Llama, respectively, over state-of-the-art (SOTA) RLVR methods, demonstrating MEML-GRPO’s ability to address the reward sparsity problem in RLVR.
\end{itemize}

\section{Related Work}\label{sec:related_work}

Recent advances in reinforcement learning (RL) have significantly enhanced the reasoning capabilities of large language models~\cite{kojima2022large,wang2022self,niu2025intent,niu2025creft}, as demonstrated by models such as DeepSeek-r1~\cite{guo2025deepseek}, OpenAI-o1~\cite{jaech2024openai}, Doubao-1.5-thinking~\cite{bytedance_seed_thinking_2025}. Notably, RL with verifiable rewards (RLVR)~\cite{wang2025reinforcement} has been systematically studied, revealing its effectiveness in enabling complex reasoning. Given the significant performance improvements achieved by RLVR, there has been growing research interest in further advancing its training procedure. For example, \citet{muennighoff2025s1,li2025s} demonstrate promising improvements in RL through test-time computation scaling, however, their experimental results reveal that performance remains constrained by the model’s inherent knowledge. \citet{ye2025limo,fatemi2025concise} demonstrate that introducing structured CoT reasoning paths can elicit advanced reasoning abilities. \citet{su2025trust,wen2025light} introduce novel training paradigms specifically designed to improve reasoning. However, recent studies~\cite{yue2025does} reveal that RLVR’s on-policy learning faces difficulties in exploration spaces, as it predominantly focuses on biasing the model toward behaviors that are more likely to yield rewards instead of learning new knowledge or reasoning paths. As a result, these methods tend to exploit familiar reasoning patterns rather than explore new knowledge-based reasoning paths. To address this challenge, we propose MEML-GRPO that leverages divers reasoning paths generated from heterogeneous pre-trained models to transcend these cognitive constraints while preserving self-driven exploration capabilities.

\section{Methodology}

The proposed framework consists of two main stages: Multi-Expert Fine-tuning (MEF) and Reinforced Inter-Expert Learning (RIEL). The former aims to equip a base language model with the ability to emulate multiple expert behaviors, while the latter further enhances the model’s performance through mutual learning among experts using reinforcement learning techniques.

\subsection{Multi-Expert Fine-tuning (MEF)}\label{sec:mef}
The MEF stage is designed to endow the base model with multi-expert capabilities by fine-tuning it on a dataset that contains answers generated by various pre-trained heterogeneous LLMs under different expert prompts. This allows the model to learn how to produce distinct responses conditioned on specific expert instructions.

\paragraph{Dataset Construction via Expert Prompting.} Let $\mathcal{E} = \{E_1, E_2, ..., E_N\} $ denote the set of $ N $ pre-trained heterogeneous expert models. For each question $Q$ in our training set $\mathcal{Q}$, we generate an answer from each expert $ E_i $:
\begin{equation}
    A^{(i)} = E_i(Q)
\end{equation}
To construct the expert-specific samples $\mathcal{D}_{\text{ME}}$, we supplement corresponding prompt after the question $Q$. This results in a multi-expert dataset:
\begin{equation}
    \mathcal{D}_{\text{ME}} = \left\{ (\texttt{Concat}(Q_j, P_i), A_j^{(i)}) \mid Q_j \in \mathcal{Q},\ i=1,...,N \right\}
\end{equation}
where $ A_j^{(i)} $ is the answer given by expert $ i $ to question $ j $. $P_i$ represents the control instruction of the $i$-th expert such as: \textit{``You are \textbf{Expert $i$}, please provide an answer to the above question."}. The function $\texttt{Concat}(\cdot)$ is used to concatenate $Q$ and $P$. A more detailed prompt would be: \textit{``You are Expert DeepSeek. [Q]''} where [Q] is the specific question. 

\paragraph{Fine-tuning Procedure.} We adopt a strong base LLM and fine-tune it using the constructed dataset $ \mathcal{D}_{\text{ME}} $. The objective is to maximize the conditional log-likelihood of the expert answers given their respective prompts:
\begin{equation}
    \mathcal{L}_{\text{MEF}} = -\sum_{j=1}^M \sum_{i=1}^N \log p_\theta \left( A_j^{(i)} \mid Q_j, \text{P}_i \right)
\end{equation}
where $ \theta $ denotes the parameters of the adopted LLM, and $ M $ is the number of questions in $ \mathcal{Q} $. After this stage, the model becomes capable of generating diverse expert-like responses depending on the input prompt. That is, for any question $ Q $, the model can be instructed to respond like expert $ i $ by simply add the corresponding prompt $P_i$ to $Q$.


   \begin{figure*}[t]
 \centering
 \includegraphics[width=1.0\linewidth]{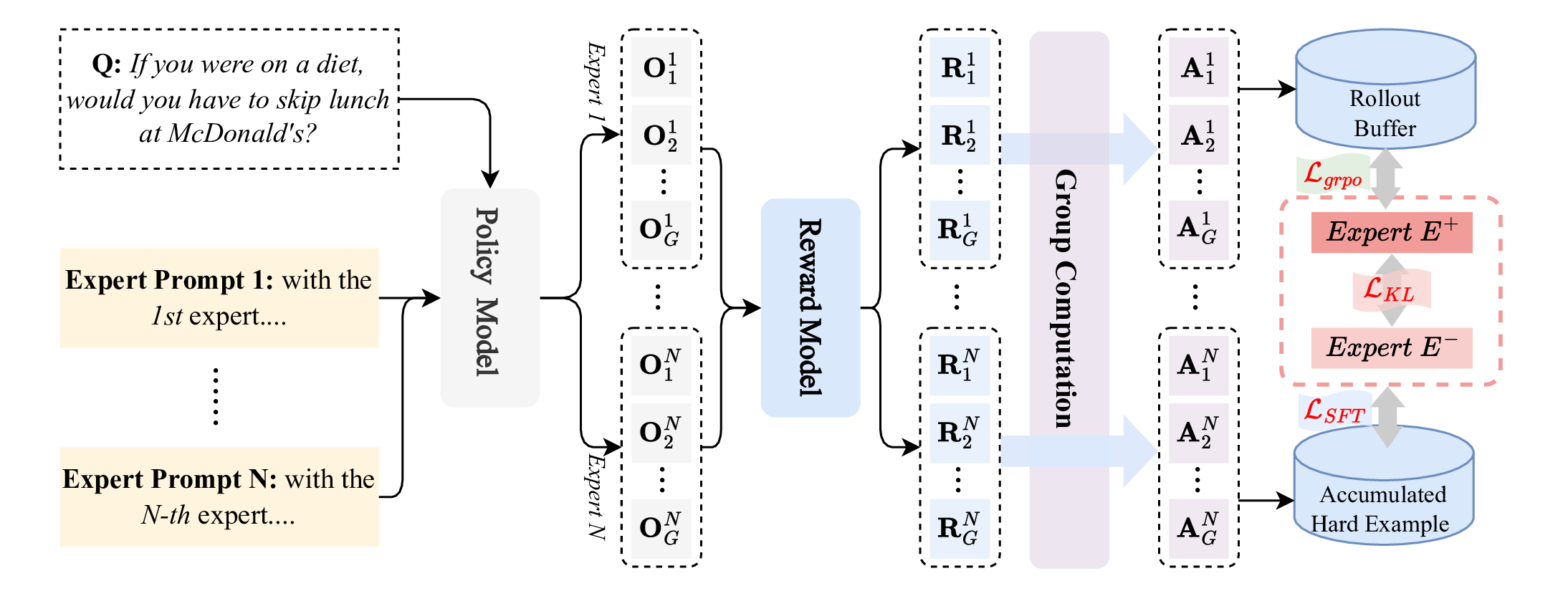}
 \caption{ This figure illustrates the pipeline of MEML-GRPO. The GRPO loss, which is computed across all experts, is omitted from the figure for brevity.
 }
 \label{framework}
 \end{figure*}

 
\subsection{Reinforced Inter-Expert Learning (RIEL)}\label{sec:riel}
While the MEF stage equips the model with expert knowledge, the RIEL stage further improves its reasoning and exploration capabilities through reinforcement learning and inter-expert knowledge transfer.

\paragraph{Response Sampling and Intra-Expert Advantage Estimation.} For a given question $ Q $, we sample $ G $ responses from each expert policy induced by the prompt $ \text{prompt}_i $:
\begin{equation}
    \mathcal{O}(Q) = \left\{ \left\{ O_1^1, ..., O_G^1 \right\}, ..., \left\{ O_1^N, ..., O_G^N \right\} \right\}
\end{equation}

Each response $O_g^i\sim\pi_{\theta}(\cdot \mid Q, P_i) $ is generated from the current policy model parameterized by $ \theta $. We then compute a reward function $ r(O_g^i) $ to evaluate the quality of each response. For example, the reward can be derived from task-specific metrics, such as accuracy, or rule-based criteria, such as matching a regular expression format.

In each expert group $i$, we compute the advantage of each response by comparing it to the average reward for that expert, following the approach in GRPO~\cite{shao2024deepseekmath}:
\begin{equation}
    A_g^i = r(O_g^i) - \frac{1}{G} \sum_{g'=1}^G r(O_{g'}^i)
\end{equation}
Using these advantages, we extend the GRPO loss for expert $ i $ as follows:
\begin{equation}
    \mathcal{L}_{\text{GRPO}}^{(i)} = -\mathbb{E}_{O_g^i \sim \pi_{\theta}} \left[ \log \pi_{\theta}(O_g^i \mid Q, P_i) \cdot \max \left( A_g^i, 0 \right) \right]
\end{equation}

The total GRPO loss across all experts is:
\begin{equation}
    \mathcal{L}_{\text{GRPO}} = \frac{1}{N} \sum_{i=1}^N \mathcal{L}_{\text{GRPO}}^{(i)}
\end{equation}
This encourages the model to reinforce responses that outperform the average within the same expert group.

\paragraph{Inter-Expert Mutual Learning via KL Divergence Regularization.} To promote knowledge exchange between experts, we introduce a mechanism called inter-expert mutual learning. For each question $Q$, we identify the best-performing expert $ E^+$ and the worst-performing expert $E^-$ based on their average reward:
\begin{equation}
\begin{aligned}
    E^+ = \arg\max_{i \in \{1, ..., N\}} \frac{1}{G} \sum_{g=1}^G r(O_g^i), \\
E^- = \arg\min_{i \in \{1, ..., N\}} \frac{1}{G} \sum_{g=1}^G r(O_g^i)
\end{aligned}  
\end{equation}

We define $prompt_{E^-}$ as the system prompt of expert $E^-$, $prompt_{E^+}$ as the system prompt of expert $E^+$. $ O^+ $ is the correct responses generated by ${E^+}$.
Thus, $ \log p_\theta(O^+\mid Q, \text{prompt}_{E^-}) $ denotes the log-probability of the high-quality response from $ E^- $, and $ \log p_\theta(O^+ \mid Q, \text{prompt}_{E^+}) $ is the log-probability of the high-quality response from $ E^+ $. 

We implement the KL divergence as a loss penalty to enable the less effective expert to learn from the more effective expert, as follows:

\begin{equation}
\begin{aligned}
    \mathcal{L}_{\text{KL}} \approx \log p_\theta(O^+ \mid Q, \text{prompt}_{E^-}) \\ - \log p_\theta(O^+ \mid Q, \text{prompt}_{E^+})
    \end{aligned}
\end{equation}

The KL divergence-based regularization term penalizes differences between the output distributions of the poorly performing expert ($E^-$) and the well-performing expert ($E^+$), encouraging the former to adopt the latter's strengths. A key benefit of this inter-expert mutual learning approach is its efficiency during inference. Unlike conventional methods like majority voting, which require multiple inferences per question (proportional to the number of experts), mutual learning enables each expert to absorb the advantages of others during training. Thus at inference time, we can select the best-performing expert, significantly reducing computational costs while maintaining high-quality responses.

\paragraph{Hard Example Accumulation via SFT Buffer.} Despite the benefits of inter-expert learning, there is a risk of error propagation when the top-performing expert also generates incorrect responses. To mitigate this issue, we design a Hard Example Accumulation Mechanism. We maintain a buffer $ \mathcal{B} $ of capacity $ B $ to store difficult samples. For each question $ Q $, if expert $ i $ produces more than $ K $ incorrect answers out of $ G $ responses, we add the pair $ (Q, P_i) \rightarrow O_{\text{gt}} $ into the buffer with probability $ \frac{K}{G} $, where $ O_{\text{gt}} $ is the correct answer. Once the buffer reaches full capacity, we perform supervised fine-tuning periodically during:
\begin{equation}
    \mathcal{L}_{\text{SFT}} = -\sum_{(Q, P_i, O_{\text{gt}}) \in \mathcal{B}} \log p_\theta(O_{\text{gt}} \mid Q, P_i)
\end{equation}

\noindent In this work, the buffer capacity (\(\mathcal{B}\)) is set to 64. A question is flagged as a hard example and added to the buffer with a probability of \(K/G = 75\%\) if experts produce more than \(K=6\) incorrect answers out of \(G=8\) samples. An SFT update is triggered only when the buffer is full. Therefore, on a challenging dataset, the buffer fills more rapidly, resulting in more frequent SFT updates. This ensures the model continues to learn from difficult cases.

\subsection{Overall Training Objective}

The overall training objective combines all components introduced above:

\begin{equation}
    \mathcal{L}_{\text{total}} = \mathcal{L}_{\text{GRPO}} + \lambda_1 \mathcal{L}_{\text{KL}} + \lambda_2 \mathcal{L}_{\text{SFT}}
\end{equation}
where $ \lambda_1 $ and $ \lambda_2 $ are hyperparameters balancing the contributions of each loss component.

\begin{table}[t]
\scriptsize
\centering
\begin{tabular}{lcccc}
\toprule
\multirow{2}{*}{\textbf{Model}} & \multicolumn{3}{c}{\textbf{Dataset}} \\
\cmidrule{2-5}
& \textbf{GSM8K} & \textbf{StrategyQA} & \textbf{MathQA} & \textbf{Avg.} \\
\midrule
\multicolumn{5}{c}{\textbf{ Qwen2.5-1.5B-Math}} \\
\midrule
Expert0-SFT & \textbf{67.0} & \textbf{69.6} & 68.0 & \textbf{68.2} \\
Expert1-SFT & 48.5 & 66.0 & 51.0 & 55.1 \\
Expert2-SFT & 66.0 & 67.2 & 65.0 & 66.0 \\
MoE-SFT (Expert0) & \textbf{67.0} & 67.0 & \textbf{70.0} & 68.0 \\
MoE-SFT (Expert1) & 48.0 & 68.0 & 53.0 & 56.3 \\
MoE-SFT (Expert2) & \textbf{67.0} & 65.0 & 65.0 & 65.6 \\
\midrule
\multicolumn{5}{c}{\textbf{Llama3.2-1B-Instruct}} \\
\midrule
Expert0-SFT & \textbf{45.0} & 55.0 & 56.0 & 52.0 \\
Expert1-SFT & 35.0 & 55.0 & \textbf{56.2} & 48.7 \\
Expert2-SFT & 42.0 & 57.0 & 48.0 & 49.0 \\
MoE-SFT (Expert0) & 42.6 & 59.3 & 54.9 & \textbf{52.2} \\
MoE-SFT (Expert1) & 36.2 & \textbf{65.8} & 46.0 & 49.3\\
MoE-SFT (Expert2) & 36.7 & 63.0 & 52.0 & 50.5\\
\bottomrule
\end{tabular}
\caption{Accuracy of MoE-SFT compared to individual Expert SFT methods. Expert0 refers to ground truth, Expert1 to DeepSeek-r1, and Expert2 to Doubao-1.5-thinking.}
\label{tab:fusing_multi_expert}
\end{table}

\begin{table*}[!h]
\centering
\footnotesize
\scalebox{0.95}{
\begin{tabular}{l p{6cm} c c}
\toprule
\textbf{Item} & \textbf{Description/Reasoning} & \textbf{Final Answer} & \textbf{Correctness} \\
\midrule
Problem & Jenny has a pizza with 12 slices. She gives \(\frac{1}{3}\) to Bill and \(\frac{1}{4}\) to Mark. After eating 2 slices herself, how many slices are left? & -- & -- \\
\midrule
Expert0 & Calculates total given: \(\frac{1}{3} + \frac{1}{4} = \frac{4}{12} + \frac{3}{12} = \frac{7}{12}\) (7 slices). Jenny eats 2 slices, leaving \(12 - 2 = 10\). Subtracts given slices: \(10 - 7 = 3\). & 3 & Yes \\
Expert1 & Gives Bill \(12 \times \frac{1}{3} = 4\) slices, Mark \(12 \times \frac{1}{4} = 3\) slices. Jenny eats 2 slices. Calculates: \(12 - 4 - 3 - 2 = 5\). & 5 & No \\
Expert2 & Calculates total given: \(\frac{1}{3} = \frac{4}{12}\), \(\frac{1}{4} = \frac{3}{12}\), so \(\frac{4}{12} + \frac{3}{12} = \frac{7}{12}\) (7 slices). Subtracts only Jenny's 2 slices: \(12 - 2 = 10\). & 10 & No \\
GT & - & 3 & -- \\
\bottomrule
\end{tabular}}
\caption{Illustration of different reasoning paths generated by different system prompts. Expert0 refers to ground truth, Expert1 to DeepSeek-r1, and Expert2 to Doubao-1.5-thinking.}\label{tab:qualitative}
\end{table*}

\section{Experimental Results}


\subsection{Experimental Settings}

\paragraph{Datasets.} We evaluate MEML-GRPO on text datasets, selecting three widely used reasoning datasets: the mathematical reasoning datasets \textbf{GSM8K}~\cite{cobbe2021training} and \textbf{MathQA}~\cite{amini-etal-2019-mathqa}, as well as the commonsense reasoning dataset \textbf{StrategyQA}~\cite{geva2021did}. Besides ground truth reasoning paths~(\textit{i.e.,} \textbf{Expert0}), other off-policy reasoning trajectories are generated by DeepSeek-r1~(\textit{i.e.,} \textbf{Expert1}), and Doubao-1.5-thinking~(\textit{i.e.,} \textbf{Expert2}).

\paragraph{Evaluation Metric.} For all datasets, as in~\citet{touvron2023llama,wang2024qwen2,qwen7b}, We use exact-match accuracy to determine correctness.

\paragraph{Baselines.} For SFT methods, we consider the following methods: (1) \textbf{Expert$X$-SFT}: models fine-tuned exclusively on specific reasoning trajectories, namely \textbf{Expert0}~(ground truth), \textbf{Expert1}~(DeepSeek-r1), and \textbf{Expert2}~(Doubao-1.5-thinking); (2) \textbf{MoE-SFT}: a single model trained on all reasoning trajectories using system prompt to distinguish reasoning style as described in Section~\ref{sec:mef}. For RLVR methods, we evaluate the following approaches: (1) \textbf{Expert$X$-GRPO}: models fine-tuned on specific reasoning trajectories and then trained using GRPO~\cite{guo2025deepseek}; (2) \textbf{Expert$X$-Dr.GRPO}: models fine-tuned on specific reasoning trajectories and then trained using Dr.GRPO~\cite{liu2025understanding}, an advanced variant of GRPO serving as a stronger baseline; (3) \textbf{MoE-SFT-GRPO}: models fine-tuned on all reasoning trajectories and then trained using GRPO; (4) \textbf{MoE-SFT-Dr.GRPO}: models fine-tuned on all reasoning trajectories and then trained using Dr.GRPO; (5) \textbf{MEML-GRPO}: models fine-tuned on all reasoning trajectories and then trained using the proposed MEML-GRPO method.

\paragraph{Training and Inference Setup.} We conduct experiments on both Qwen2.5-1.5B-Math~\cite{yang2024qwen2}, Llama3.2-1B-Instruct~\cite{grattafiori2024llama} to verify the generalization of MEML-GRPO. To ensure fairness, we maintain 8 rollouts per prompt for all RL-trained models. The learning rate is set to $1\times10^{-6}$. All training experiments are conducted on 8 A800 GPUs. SFT training setup, The learning
rate is set to $1\times10^{-5}$. We train all RL models for 1 epoch and all SFT models for 1 epochs. For inference, to eliminate the impact of randomness, no sampling methods are employed during testing for any of the models. Greedy search is used for generation across all models.

\subsection{Main Results}

\begin{figure}[!ht]
\centering
\begin{subfigure}{\columnwidth}
    \includegraphics[width=\columnwidth]{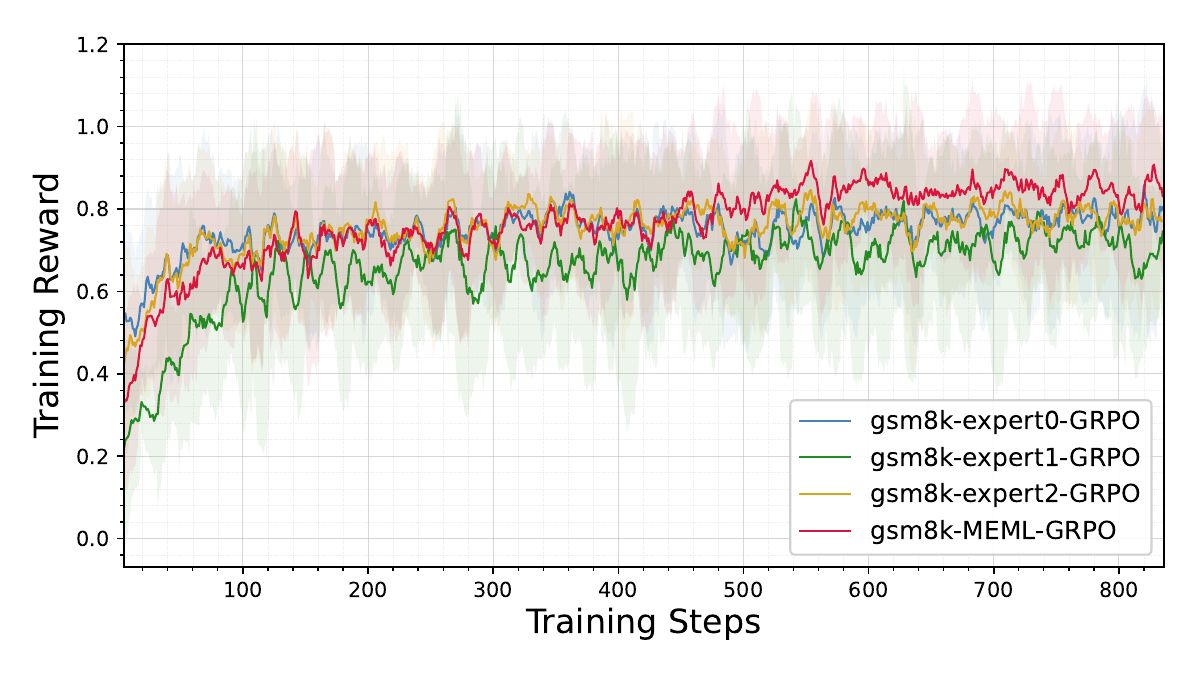}
    \caption{GSM8K}
    \label{subfig:reward1}
\end{subfigure}
\hfill
\begin{subfigure}{\columnwidth}
    \includegraphics[width=\columnwidth]{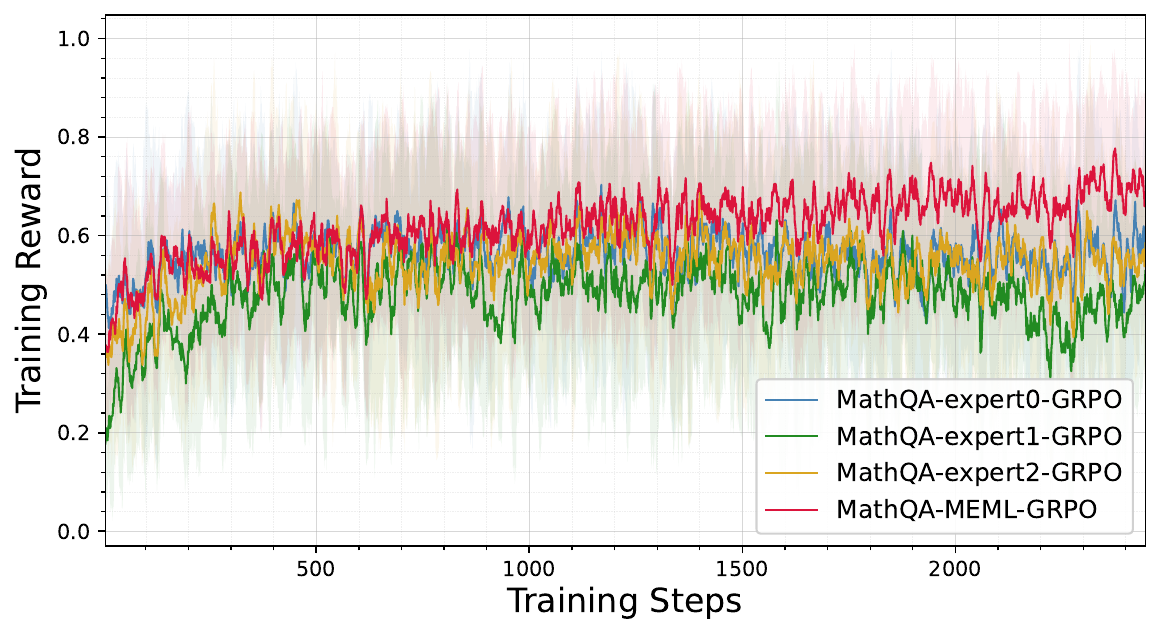}
    \caption{MathQA}
    \label{subfig:reward2}
\end{subfigure}
\hfill
\begin{subfigure}{\columnwidth}
    \includegraphics[width=\columnwidth]{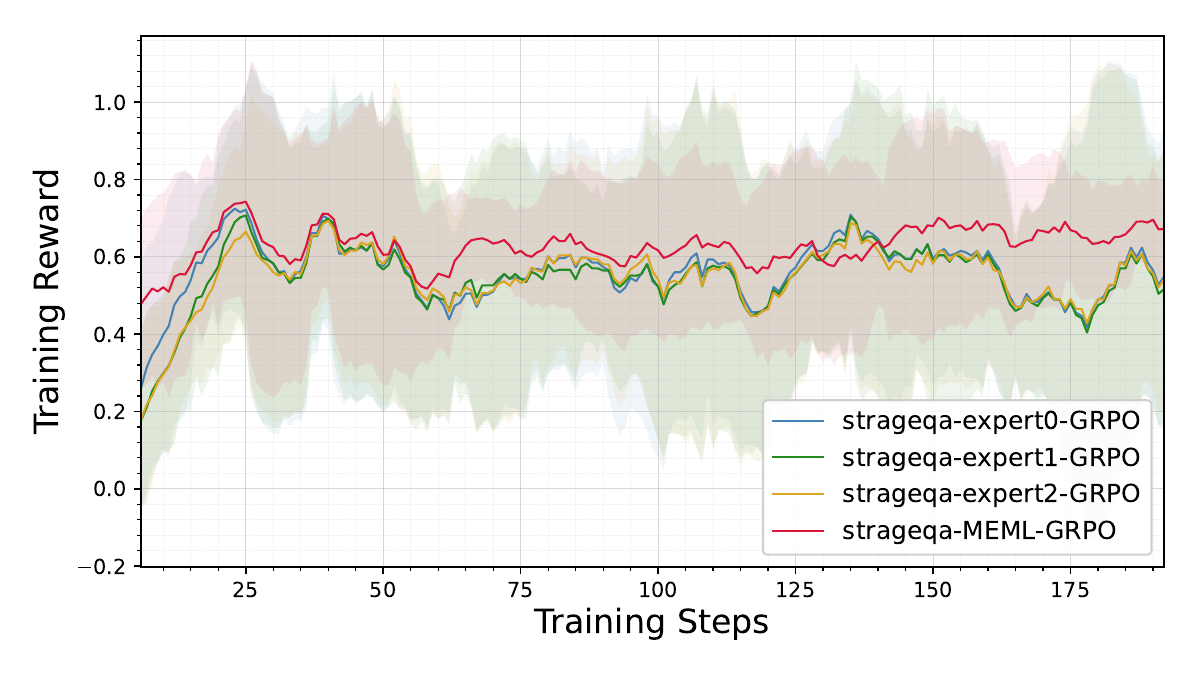}
    \caption{StrategyQA}
    \label{subfig:reward3}
\end{subfigure}
\caption{Training reward dynamics of MEML-GRPO (Llama3.2) compared with other on-policy RL.}
\label{fig:train_dynamics}
\end{figure}

\begin{table}[!ht]
\scriptsize
\centering
\scalebox{0.95}{
\begin{tabular}{lcccc}
\toprule
\multirow{2}{*}{\textbf{Model}} & \multicolumn{4}{c}{\textbf{Dataset}} \\
\cmidrule{2-5}
& \textbf{GSM8K} & \textbf{StrategyQA} & \textbf{MathQA} & \textbf{Avg.} \\
\midrule
\multicolumn{5}{c}{\textbf{ Qwen2.5-1.5B-Math}} \\
\midrule
\multicolumn{5}{c}{\textbf{ Single Expert}} \\
\midrule
Expert0-SFT-GRPO & 75.9 & 58.6 & 71.3 & 67.1 \\
Expert1-SFT-GRPO & 73.1 & 62.5 & 71.8 & 69.1 \\
Expert2-SFT-GRPO & 76.2 & 60.4 & 69.3 & 68.6 \\
Expert0-SFT-Dr.GRPO & 78.3 & 55.2 & 72.4 & 68.6 \\
Expert1-SFT-Dr.GRPO & 76.1 & 64.5 & 73.3 & 71.3 \\
Expert2-SFT-Dr.GRPO & 77.2 & 62.4 & 70.6 & 70.0 \\
\midrule
\multicolumn{5}{c}{\textbf{ Multiple Expert}} \\
\midrule
MoE-SFT-GRPO (Expert0) & 77.1 & 68.2 & 70.4 & 71.9\\
MoE-SFT-GRPO (Expert1) & 75.6 & 70.3 & 68.5 & 71.4 \\
MoE-SFT-GRPO (Expert2) & 74.3 & 67.9 & 70.4 & 70.8 \\
MoE-SFT-Dr.GRPO (Expert0) & 77.3 & 68.6 & 72.1 & 72.6 \\
MoE-SFT-Dr.GRPO (Expert1) & 77.7 & 69.3 & 74.2 & 73.7 \\
MoE-SFT-Dr.GRPO (Expert2) & 77.6 & 68.7 & 70.1 & 72.1 \\
\midrule
\multicolumn{5}{c}{\textbf{ Ours}} \\
\midrule
MEML-GRPO (Expert0) & \textbf{80.1} & 73.1 & \textbf{76.4} & 76.5 \\
MEML-GRPO (Expert1) & 79.6 & \textbf{75.3} & \textbf{76.4} & \textbf{77.1} \\
MEML-GRPO (Expert2) & 78.1 & 74.1 & 76.3 & 76.2 \\
\midrule
\multicolumn{5}{c}{\textbf{Llama3.2-1B-Instruct}} \\
\midrule
\multicolumn{5}{c}{\textbf{ Single Expert}} \\
\midrule
Expert0-SFT-GRPO & 56.3 & 54.1 & 52.5 & 54.3 \\
Expert1-SFT-GRPO & 58.2 & 54.0 & 45.2 & 52.4 \\
Expert2-SFT-GRPO & 54.5 & 53.1 & 47.9 & 51.8 \\
Expert0-SFT-Dr.GRPO & 57.3 & 55.6 & 54.3 & 55.7 \\
Expert1-SFT-Dr.GRPO & 58.4 & 54.8 & 46.7 & 53.3 \\
Expert2-SFT-Dr.GRPO & 54.5 & 53.7 & 48.5 & 52.2 \\
\midrule
\multicolumn{5}{c}{\textbf{ Multiple Expert}} \\
\midrule
MoE-SFT-GRPO (Expert0) & 57.7 & 53.7 & 55.1 & 55.5 \\
MoE-SFT-GRPO (Expert1) & 57.1 & 53.8 & 55.2 & 55.3 \\
MoE-SFT-GRPO (Expert2) & 56.7 & 53.4 & 54.0 & 54.7 \\
MoE-SFT-Dr.GRPO (Expert0) & 57.9 & 53.1 & 54.6 & 55.2 \\
MoE-SFT-Dr.GRPO (Expert1) & 57.2 & 54.2 & 54.2 & 55.2 \\
MoE-SFT-Dr.GRPO (Expert2) & 57.4 & 53.1 & 53.9 & 54.8 \\
\midrule
\multicolumn{5}{c}{\textbf{ Ours}} \\
\midrule
MEML-GRPO (Expert0) & 60.3 & 62.0 & \textbf{60.8} & \textbf{61.1} \\
MEML-GRPO (Expert1) & 61.0 & \textbf{63.3} & 58.4 & 60.9 \\
MEML-GRPO (Expert2) & \textbf{61.3} & 61.7 & 58.8 & 60.6 \\
\bottomrule
\end{tabular}}
\caption{Accuracy of MEML-GRPO compared to other RL baselines. Expert0 refers to ground truth, Expert1 to DeepSeek-r1, and Expert2 to Doubao-1.5-thinking.}
\label{tab:fusing_multi_expert_rl}
\end{table}

\textbf{Multi-expert Learning for SFT.} As presented in Table~\ref{tab:fusing_multi_expert}, for instance, in the GSM8K dataset with Llama3.2 the three experts in MoE-SFT achieve scores of 42.6\%, 36.2\%, and 36.7\%, respectively. These results are slightly lower than those of individually trained Expert0-SFT (45.0\%) and Expert2-SFT (42.0\%). A similar trend is observed in MathQA and StrategyQA, as well as with Qwen2.5. The notable exception is StrategyQA with Llama3.2 where one expert achieves a score of 65.8\%, significantly surpassing the performance of individually trained SFT experts. These findings suggest that MoE-SFT can effectively leverage the knowledge of multiple experts to some degree, with each system prompt enabling the same model to function as distinct experts, reflecting the reasoning behaviors of corresponding heterogeneous models. However, the absence of an effective mechanism to fully utilize the knowledge of multiple experts limits MoE-SFT's ability to consistently outperform individually trained expert SFT models across all scenarios, highlighting the effectiveness of MEML-GRPO.

\begin{table*}[t]
\centering
\scriptsize
\begin{tabular}{ccc|cccc|cccc}
\toprule
~&~& ~&\multicolumn{4}{c|}{\textbf{Qwen2.5-1.5B-Math}} & \multicolumn{4}{c}{\textbf{Llama3.2-1.5B-Instruct}} \\\midrule
  MoE & HSFT &  IML & GSM8K & StrategyQA & MathQA  & Avg &GSM8K & StrategyQA & MathQA & Avg  \\    \midrule
 $\times$  & $\times$   & $\times$ & 76.2 & 62.5 & 71.8 & 70.1 & 56.3 & 54.1 &  52.5 & 54.3\\
 \checkmark & $\times$  & $\times$  & 77.1 & 70.3 & 70.4 & 72.6 & 57.7 & 53.8 & 55.2 & 55.5 \\
 \checkmark & \checkmark  & $\times$ & 79.9 & 74.3 & 75.1 &  76.4 & 59.4 & 61.2  & 58.8 & 59.8 \\
 \checkmark & $\times$  &\checkmark & 78.1 & 72.3 & 73.4 & 74.6  & 58.4 & 56.8  & 57.3 & 57.5 \\
 \checkmark & \checkmark &  \checkmark & \textbf{80.1} & \textbf{75.3}  & \textbf{76.4} & \textbf{77.3} & \textbf{61.3} & \textbf{63.3} & \textbf{60.8} & \textbf{61.8}  \\

\bottomrule
\end{tabular}
\caption{Ablations on each component of MEML-GRPO. For brevity, we report the results of best-performing experts. MoE: multiple expert SFT. HSFT: hard example SFT. IML: Inter-expert mutual learning.}
\label{tab:ablation_study}
\end{table*}

\begin{table}[!ht]
\scriptsize
\centering
\begin{tabular}{lcccc}
\toprule
\multirow{2}{*}{\textbf{Model}} & \multicolumn{4}{c}{\textbf{Dataset}} \\
\cmidrule{2-5}
& \textbf{GSM8K} & \textbf{StrategyQA} & \textbf{MathQA} & \textbf{Avg.} \\
\midrule
\multicolumn{5}{c}{\textbf{ Qwen2.5-1.5B-Math}} \\
\midrule
\multicolumn{5}{c}{\textbf{ Multiple Expert}} \\
\midrule
MoE-SFT-GRPO (Expert0) & 77.1 & 68.2 & 70.4 & 71.9 \\
MoE-SFT-GRPO (Expert1) & 75.6 & 70.3 & 68.5 & 71.4 \\
MoE-SFT-GRPO (Expert2) & 74.3 & 67.9 & 70.4 & 70.8 \\
MoE-SFT-GRPO (MV) & \textbf{78.3} & \textbf{71.1} & \textbf{71.6} & \textbf{73.6} \\
Delta & 1.2 & 0.8 & 1.2 & 1.1 \\
\midrule
\multicolumn{5}{c}{\textbf{ Ours}} \\
\midrule
MEML-GRPO (Expert0) & \textbf{80.1} & 73.1 & \textbf{76.4} & 76.6 \\
MEML-GRPO (Expert1) & 79.6 & \textbf{75.3} & \textbf{76.4} & \textbf{77.1} \\
MEML-GRPO (Expert2) & 78.1 & 74.1 & 76.3 & 76.2 \\
MEML-GRPO (MV) & 79.8 & 75.1 & 75.9 & 76.9 \\
Delta & -0.3 & -0.2 & -0.5 & -0.3 \\
\bottomrule
\end{tabular}
\caption{Comparison of majority voting performance versus a single expert, with delta calculated as the difference between majority vote performance and the best-performing expert. MV: majority voting.}
\label{tab:majority_vote}
\end{table}

\paragraph{Comparison with SOTA RLVR methods.} Table~\ref{tab:fusing_multi_expert_rl} compares the performance of MEML-GRPO with other RL baselines. MEML-GRPO consistently achieves top performance across all datasets for both Qwen2.5 and Llama3.2 models. Note that to ensure a fair comparative evaluation, we partition the dataset into two distinct subsets: 20\% for warm-up SFT and the remaining 80\% for RL optimization, differing from the setup used in Table~\ref{tab:fusing_multi_expert}. 
We first compare MEML-GRPO with the single-expert RL method, Expert0-SFT-GRPO, which serves as the standard GRPO baseline since Expert0 is fine-tuned with ground-truth reasoning data. Using Qwen2.5, MEML-GRPO(Expert0) outperforms this baseline by significant margins of 4.2\%, 14.5\%, and 5.1\% across the datasets. Similarly, when compared to Expert0-SFT-Dr.GRPO, a standard Dr.GRPO method, MEML-GRPO demonstrates substantial improvements. Consistent trends are observed with Llama3.2, highlighting the robustness of MEML-GRPO across different models. Note that the single-expert RL method, when trained with data generated by other experts, also performs less effectively than the MEML-GRPO with the corresponding expert. 


\paragraph{Comparison with Other Multi-expert RLVR Methods.} 

By integrating responses from all experts in the training set, standard GRPO and Dr.GRPO can be extended to Multi-expert RLVR methods, namely MoE-SFT-GRPO and MoE-SFT-Dr.GRPO, establishing stronger baselines. Table~\ref{tab:fusing_multi_expert_rl} shows that the MEML-GRPO method substantially outperforms state-of-the-art approaches across all reasoning tasks, delivering consistent improvements with both Qwen2.5 and Llama3.2 models, as evidenced by its superior overall performance. For example, on the Qwen2.5-1.5B-Math model, MEML-GRPO achieves an average accuracy of 79.3\% on the GSM8K dataset, surpassing its GRPO and Dr.GRPO counterparts. Notably, on StrategyQA, the three trained experts under MEML-GRPO achieve accuracies of 73.1\%, 75.3\%, and 74.1\%, substantially outperforming GRPO (68.2\%, 70.3\%, 67.9\%) and Dr.GRPO (68.6\%, 69.3\%, 68.7\%). Similarly, with Llama3.2, MEML-GRPO also demonstrates superior performance across all three datasets. 

The possible reasons for the performance differences mainly lie in the two core designs of MEML-GRPO: (1) The multi-expert prompt mechanism generates more diverse responses through varied system prompts, significantly increasing the probability of covering correct solutions. This particularly alleviates the reward sparsity problem in traditional RLVR where ``no learning signal is provided when all candidate answers are wrong'' in complex reasoning tasks. (2) The inter-expert mutual learning mechanism facilitates knowledge sharing and transfer among experts, further breaking through the capability ceiling of the model through reinforcement learning. In contrast, MoE-SFT-GRPO and MoE-SFT-Dr.GRPO only rely on a single reinforcement learning strategy. They lack active enhancement of response diversity and collaborative learning among experts, thus failing to fully exploit the model's potential in reasoning tasks and resulting in limited performance. 

Table~\ref{tab:qualitative} presents a qualitative comparison of reasoning paths provided by different experts~(\textit{i.e.,} system prompts) for a pizza slice distribution problem that is sampled from GSM8K. Figure~\ref{fig:train_dynamics} shows the training reward dynamics of MEML-GRPO~(Llama3.2) compared with other on-policy RL method, the results with Qwen2.5 are in Appendix.

\subsection{Discussion}

\paragraph{On the Benifit of Mutual Learning.} Table~\ref{tab:majority_vote} presents the distinct differences in delta values, demonstrating that our MEML-GRPO models outperform the majority vote (MV) approach across all datasets. Compared to majority voting, MEML-GRPO achieves higher accuracy, as evidenced by the delta values. This improvement highlights the effectiveness of mutual learning in transferring knowledge between different experts, enabling the reasoning results to rival or even match the performance of multi-expert majority voting.
In contrast, conventional schemes typically achieve their best results with majority voting. The results with Llama3.2 can be found in Appendix.

\paragraph{Ablation Study.} 


Table~\ref{tab:ablation_study} shows the impact of individual components in the MEML-GRPO framework. The baseline achieved average scores of 70.1\% for Qwen2.5 and 54.3\% for Llama3.2. Enabling Mixture of Experts (MoE) alone improved the averages to 72.6\% and 55.5\%, respectively. Adding Hypothesis Selection Fine-Tuning (HSFT) further increased the scores to 76.4\% and 59.8\%. Incorporating Interactive Multi-step Learning (IML) alone provided an average gain of ~2\% for both models. The full configuration (MoE+HSFT+IML) yielded the highest averages: 77.3\% for Qwen2.5 and 61.8\% for Llama3.2, showing that each component contributes and their combination is most effective.

\paragraph{Training and Inference Cost}

MEML-GRPO achieves remarkable training efficiency, with a total cost significantly lower than expected. By utilizing the vLLM inference engine for parallel batched rollouts and paged attention, the total training time increases by only 20-30\% over the single-expert baseline—far less than the theoretical N-fold increase (where N=3, the number of experts). Furthermore, MEML-GRPO's peak memory footprint is comparable to conventional GRPO (around 60GB), as vLLM pre-allocates memory based on inference settings. This efficiency extends to inference: after training, only a single model is deployed. By dynamically selecting the best expert prompt, MEML-GRPO delivers superior performance with the latency of a single model, avoiding the computational overhead of ensemble methods that require running N models per input.

\section{Conclusion}

In this work, we introduce MEML-GRPO, a multi-expert mutual learning framework that mitigates reward sparsity in RL for LLM reasoning by leveraging complementary strengths of diverse pre-trained models. Experiments on GSM8K, MathQA, and StrategyQA show consistent improvements over SOTA RLVR methods, with ablations confirming the contribution of each component.


\bibliography{aaai2026}

\end{document}